\documentclass{article}
\usepackage{ijcai24}
\usepackage{xcolor}
\usepackage{multirow}
\usepackage{times}
\usepackage{soul}
\usepackage{url}
\usepackage[hidelinks]{hyperref}
\usepackage[utf8]{inputenc}
\usepackage[small]{caption}
\usepackage{graphicx}
\usepackage{amsmath}
\usepackage{amsthm}
\usepackage{booktabs}
\usepackage{algorithm}
\usepackage{algorithmic}
\usepackage[switch]{lineno}

\usepackage{amssymb}

\title{HFedCKD: Toward Robust Heterogeneous Federated Learning via Data-free Knowledge Distillation and Two-way Contrast}

\author{
    Yiting Zheng\thanks{These authors contributed equally and are co-first authors.}\and
    Bohan Lin\footnotemark\and
    Jinqian Chen\and
    Jihua Zhu\thanks{Corresponding Author}\\
    \affiliations
    School of Software Engineering, Xi’an Jiaotong University\\
    \emails
    \{zhengyiting, linbohan, chenjinqian\}@stu.xjtu.edu.cn, zhujh@xjtu.edu.cn
}

\begin{document}

\maketitle

\begin{abstract}
    Most current federated learning frameworks are modeled as static processes, ignoring the dynamic characteristics of the learning system. Under the limited communication budget of the central server, the flexible model architecture of a large number of clients participating in knowledge transfer requires a lower participation rate, active clients have uneven contributions, and the client scale seriously hinders the performance of FL. We consider a more general and practical federation scenario and propose a system heterogeneous federation method based on data-free knowledge distillation and two-way contrast (HFedCKD). We apply the Inverse Probability Weighted Distillation (IPWD) strategy to the data-free knowledge transfer framework. The generator completes the data features of the nonparticipating clients. IPWD implements a dynamic evaluation of the prediction contribution of each client under different data distributions. Based on the antibiased weighting of its prediction loss, the weight distribution of each client is effectively adjusted to fairly integrate the knowledge of participating clients. At the same time, the local model is split into a feature extractor and a classifier. Through differential contrast learning, the feature extractor is aligned with the global model in the feature space, while the classifier maintains personalized decision-making capabilities. HFedCKD effectively alleviates the knowledge offset caused by a low participation rate under data-free knowledge distillation and improves the performance and stability of the model. We conduct extensive experiments on image and IoT datasets to comprehensively evaluate and verify the generalization and robustness of the proposed HFedCKD framework.

\end{abstract}

\section{Introduction}

 Federated learning (FL) has broad application prospects in theory, but existing research methods often ignore practical application conditions. In an actual IoT environment, there are significant differences in the computing resources, model architecture, and communication bandwidth of devices \cite{zhang2021federated}. FL faces the following two key challenges in actual implementation: (1) Model heterogeneity. In actual applications, different clients may deploy different model architectures, such as ResNet, MobileNet, and MLP. parameter structures of these heterogeneous models cannot be directly aggregated, making it difficult for traditional FL methods (such as FedAvg) and their variants to adapt to diverse network model architectures \cite{mcmahan2017communication,Peng2024FedPFTFP}. (2) Limited communication bandwidth. With the explosive growth of the number of Internet users, the number of clients participating in FL continues to increase, but the limitation of communication bandwidth means that the participation rate of the clients must be reduced\cite{Shi2020JointDS,10.1109/JSAC.2024.3431516,Liu2020}. The low participation rate further exacerbates the imbalance in the distribution of client data, resulting in a significant decrease in global model performance.

\begin{figure}[t]
	\centering
        \includegraphics[width=\columnwidth]{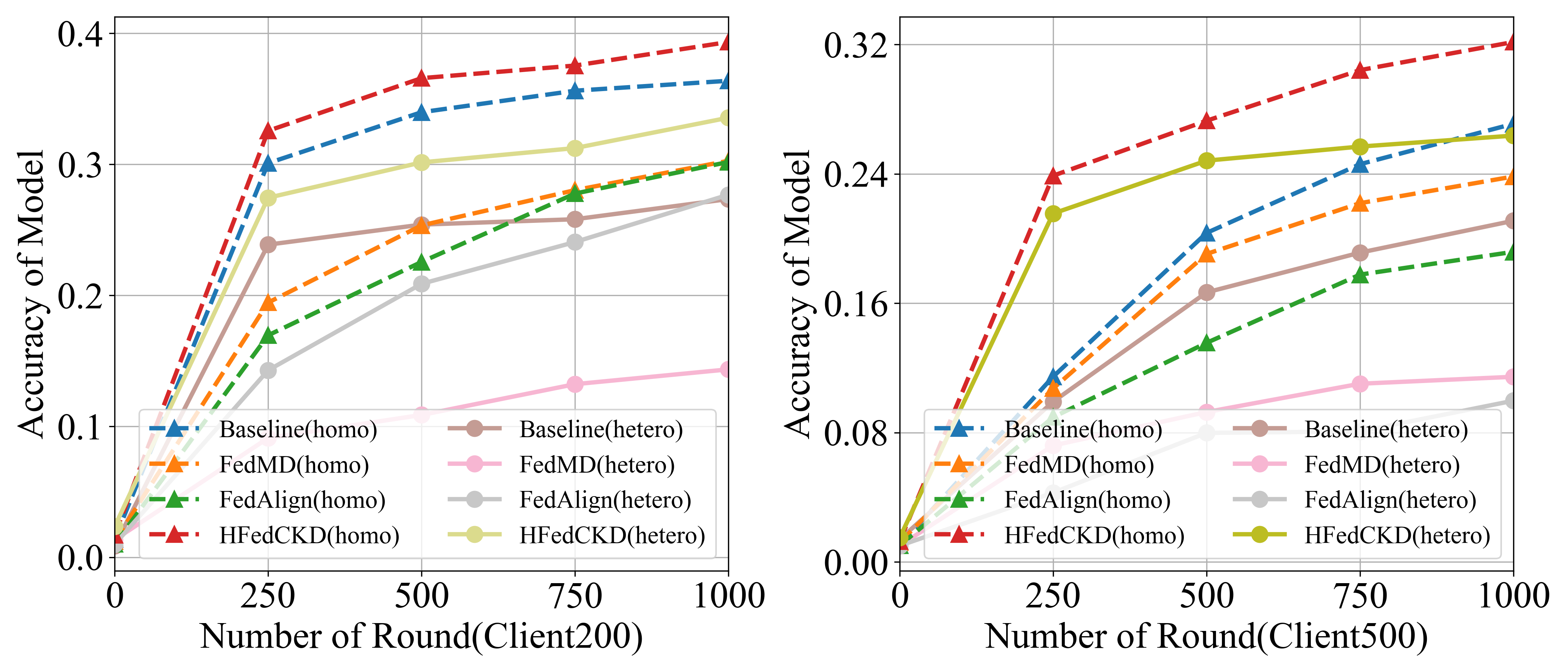}
	\caption{The comparison of performance between homogeneous and heterogeneous models on the CIFAR-100 dataset with 200 clients (left) and 500 clients (right).} 
        \label{fig:Figure 1}
\end{figure}

In response to model heterogeneity, there are currently two types of methods that have been attempted: one is based on Partial Training (PT), which only supports client models to achieve narrow-sense training by scaling shared architectures (such as submodels of the same model) \cite{alam2022fedrolex,10.5555/3666122.3666906,lee2022preservation,diao2021heterofl}. The heterogeneity of the model limits its applicability in complex actual scenarios; the other type is methods based on Knowledge Distillation (KD), which usually require high-quality public data but face high privacy risks \cite{wu2024exploring,10.1145/3638252,venkateswaran2023fedgen,mendieta2022local}. These issues have become core obstacles that hinder the widespread application of FL in privacy-preserving and resource-constrained scenarios. At the same time, the adaptability of the above methods under limited communication bandwidth conditions has not been fully studied. As the scale of IoT users expands further, limitations in the communication bandwidth will pose more severe challenges to FL performance.

To address this issue, we conducted an in-depth study on the actual impact of communication bandwidth in scenarios with heterogeneous data and models. We find that under limited communication bandwidth conditions, the local model performance decreases significantly as the participation rate decreases. This problem is more prominent in model heterogeneous scenarios than in homogeneous scenarios.

Based on the above findings, we propose a novel system structure framework (HFedCKD) based on data-free knowledge supplementation and structural contrast learning, aiming at the sudden explosion and false samples that appear in the generator under low participation rate conditions in data-free knowledge supplementation \cite{reddi2021adaptive,Wu2021FedCGLC,10.1007/978-3-031-39698-4_23}. To deal with the quality degradation problem, we introduce an IPWD to give different weights to the client's participation frequency and data quality through a dynamic allocation strategy \cite{niu2022respecting}. After the weight of low-frequency participating clients is increased, the generator can capture more even global distribution information and reduce dependence on single client data, thus improving the stability of the generator.At the same time, we set up Hierarchical Contrastive Learning (HCL) to decompose the model into Encode and Decode \cite{feng2023hierarchical,collins2021exploiting}. Encode-Global Alignment: By aligning the features output by the generator with the global model feature space, it reduces the difference in feature distribution between clients and alleviates label deviation and feature drift problems; Decode-History Alignment: By introducing feature alignment of historical models, Preserve the client's personalized optimization path to ensure that the local model does not deviate from its data characteristics during the update process\cite{conf/aaai/ChenZZLT24}. This two-way optimization mechanism not only achieves global alignment, but also protects the diversity of local models, greatly improving the model's generalization ability and local performance in Non-IID data scenarios\cite{10.1145/3704323.3704370}. We conducted comparative experiments on three classic datasets, and the results show that the proposed HFedCKD framework still has excellent robustness and performance advantages under low participation rate settings. Main contributions:

\begin{itemize}
\item We take the first attempt to explore the impact of limited communication bandwidth in practical application scenarios with heterogeneous models and data. We found that low participation rate will significantly deteriorate model performance, especially in model heterogeneous scenarios, which will have a more serious impact on the convergence and accuracy of the model.

\item  HFedCKD introduces IPWD and sets up network layered differential contrastive learning. IPWD alleviates generator bias caused by uneven data contributions. The decoder and compiler conduct two-way comparative learning to enhance the global model feature expression ability and the client's personalized decision-making ability. HFedCKD effectively combats label bias and drift caused by low participation rates.

\item We completed experimental verification on multiple tasks on image and IoT datasets to comprehensively evaluate the generalization and robustness of the proposed heterogeneous federated learning(HFL) framework. Experimental results show that the HFedCKD framework can effectively resist the damage of low participation rates in heterogeneous scenarios, achieve efficient collaboration with diverse clients, and is more in line with actual implementation needs.
\end{itemize}

\section{Related Work}

\subsection{Heterogeneous Federated Learning}
Most FL methods are based on the idealized assumption that all clients' local models have the same architecture as the global model. This premise limits the application of these methods in actual production environments. In this regard, the HeteroFL method supports incomplete model heterogeneity between clients through weighted aggregation and local model calibration, and its applicability is limited to client models that are scaled differently \cite{diao2021heterofl}.FedRolex has performed performance optimization on the basis of HeteroFL, but the aggregation strategy is still difficult to adapt to complete model heterogeneity \cite{alam2022fedrolex}. FjORD\cite{horvath2021fjord} supports incomplete heterogeneity by training only part of the hierarchy of the model. FedGKT and FedDF replace traditional model parameter aggregation through knowledge distillation, and the server side has high-quality proxy datasets \cite{he2020group,zhu2021federated}. Although DSFL supports model heterogeneity through knowledge distillation, it ignores the refined distillation of internal features of the model, thus affecting the performance of the model \cite{10019204}. These also ignore the limitation and volatility of communication bandwidth in dynamic network environments. In low participation rates or bandwidth-limited scenarios, these methods may cause model training to fail or even fail.

\subsection{Inverse Probability Weighting Distillation}
IPWD is an improved method to address the delivery gap problem that exists in KD. The traditional KD assumes that the training samples of the teacher model and the student model are IID in the human field and the machine field, but this assumption is difficult to meet the actual situation, especially in the knowledge transfer of a small number of classes \cite{chen2023towards,usmanova2022federated}. IPWD compensates for this imbalance by assigning a weight to each training sample by estimating the propensity score of each sample and inversely assigning it to the sample weight. Undervalued samples are given greater weight, so that the knowledge of these samples can be better transferred to the student model.

\subsection{Data-free Knowledge Distillation}
Data-free knowledge distillation aims to effectively transfer global knowledge to heterogeneous client models through the mechanism of knowledge distillation without relying on shared public data or actively generating data. FedGen generates public knowledge through a server generator, and the client uses this knowledge to distill local models, but its performance is limited by the generator quality \cite{venkateswaran2023fedgen}. DFRD combines data-free knowledge distillation with a teacher-student model architecture to achieve collaborative training of heterogeneous models \cite{10.5555/3666122.3666906}. When communication bandwidth is limited, it is difficult for the generator to obtain sufficient and high-quality client feedback, resulting in incomplete coverage of the generated public knowledge.

\subsection{Contrastive Learning}
The contrastive learning method in FL gradually improves the adaptability and robustness of the model under Non-IID data distribution by combining the feature extraction advantages of contrastive learning. FedCL extracts features through comparative learning locally on the client, and improves the model's adaptability to Non-IID data distribution through global aggregation \cite{10.1145/3639706}; MP-FedCL captures the diversity of data distribution through a multi-prototype mechanism to further enhance the adaptability of Non-IID The robustness of the data enables more efficient federated comparative learning \cite{10.5555/3692070.3692967}; MOON optimizes the representation consistency between the global model and the local model through comparative learning, thereby effectively alleviating the impact of inconsistent data distribution on model performance \cite{li2021model}.These methods jointly promote the performance improvement of FL in heterogeneous scenes. However, the reliance on negative sample selection can lead to high communication overhead for feature representation, which shows certain limitations when processing complex multi-modal data\cite{10.24963/ijcai.2024/419}.


\begin{figure}[t]
	\centering
        \includegraphics[height=5.5cm]{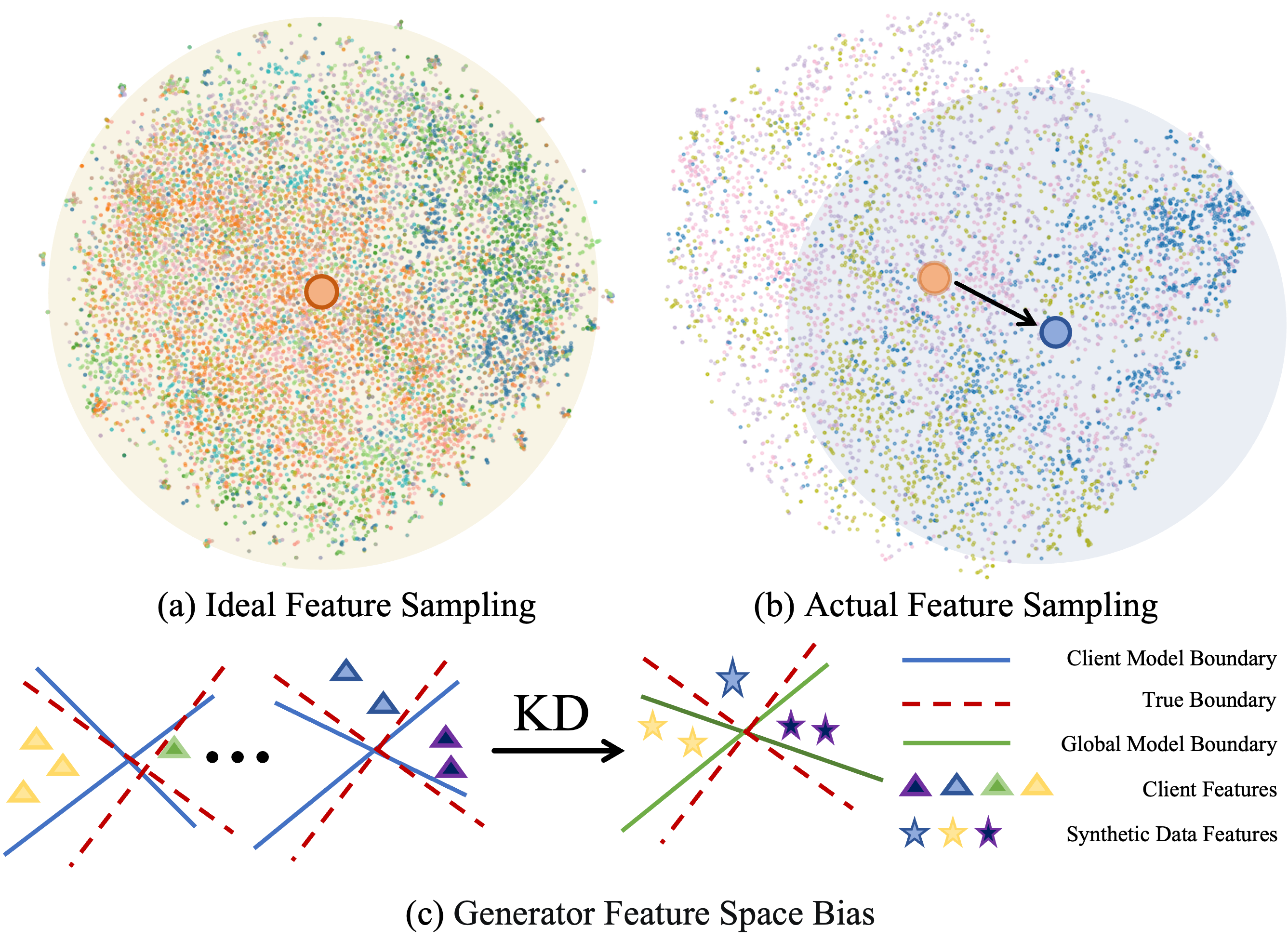}
	\caption{In Fig.2a and 2d, we project the feature representation onto the 2D plane by t-SNE. The goal of the global model is to learn the knowledge of the client. When updated in each round, the generated model should be able to effectively capture the characteristics of each client. However, in fact, there is a large deviation in feature distribution between different clients. A low participation rate will cause the update of the global model to only reflect the feature distribution of participating clients, causing a bias in the global optimization goal. In Fig.2c, low participation rate makes the generator unable to obtain information from the features of non-participating clients, the quality of the generated pseudo samples decreases, and the feature diversity is lost, causing the update of the global model to deviate from the direction of global optimization.} 
        \label{fig:Figure 1}
\end{figure}

\section{Motivation}
Low participation rate will amplify the impact of model heterogeneity and data heterogeneity, which is essentially insufficient data distribution coverage and reduced feature expression ability.In scenarios with large differences in data distribution and model architecture, when the client participation rate is low, the generator tends to learn the distribution characteristics of high-participation clients and ignore the data characteristics of low-participation clients, making it difficult to optimize the generator and easily falling into local optimality, further weakening the effect of knowledge transfer, thus exacerbating the imbalance of global knowledge;
\paragraph{Impact on generator.}Suppose there are $N$ clients in a federated learning system, and the number of clients participating in each round is $C$ ($C \ll N$).
The data-free knowledge distillation method relies on pseudo samples generated by the generator $G$ to simulate the global data distribution. The goal of the generator is to generate a sample distribution $P_G(x)$ that is close to the real data distribution $P(x)$. When the participation rate is low (i.e., $|\mathcal{C}_t| \ll N$), only the local data $D_i \sim P_i(x)$ of the participating clients is used to train the generator, $P(x)$ is replaced by the partial distribution $\{P_i(x)\}_{i \in \mathcal{C}_t}$, and the generator cannot fully capture the feature distribution of the non-participating clients, causing $P_G(x)$ to deviate from $P(x)$. For a certain category $c$, the data of the non-participating clients may contain unique features of the category. Under low participation rate conditions, the samples generated by the generator lack these features, resulting in the deviation of the pseudo sample distribution $P_G(x \mid c)$ from the true distribution $P_{\text{global}}(x \mid c)$:


\begin{equation}
     D_{{KL}}(P_G(x \mid c) \| P(x \mid c)) \to \infty \quad 
\end{equation}

This causes the quality of pseudo samples to deteriorate and feature diversity to be lost.

\paragraph{Impact on the global model.}In the ideal case where all clients participate in training, the knowledge distillation loss $\mathcal{L}_{\text{KD}, i}$ of each client contributes evenly to the global objective function: 
\begin{equation}
F(\mathbf{w}_{\text{global}}) = \frac{1}{N} \sum_{i=1}^N \mathcal{L}_{\text{KD}, i},
\end{equation}
where $\mathbf{w}_{\text{global}}$ is global model parameters;

\begin{figure*}[t!]
	\centering
        \includegraphics[height=6cm]{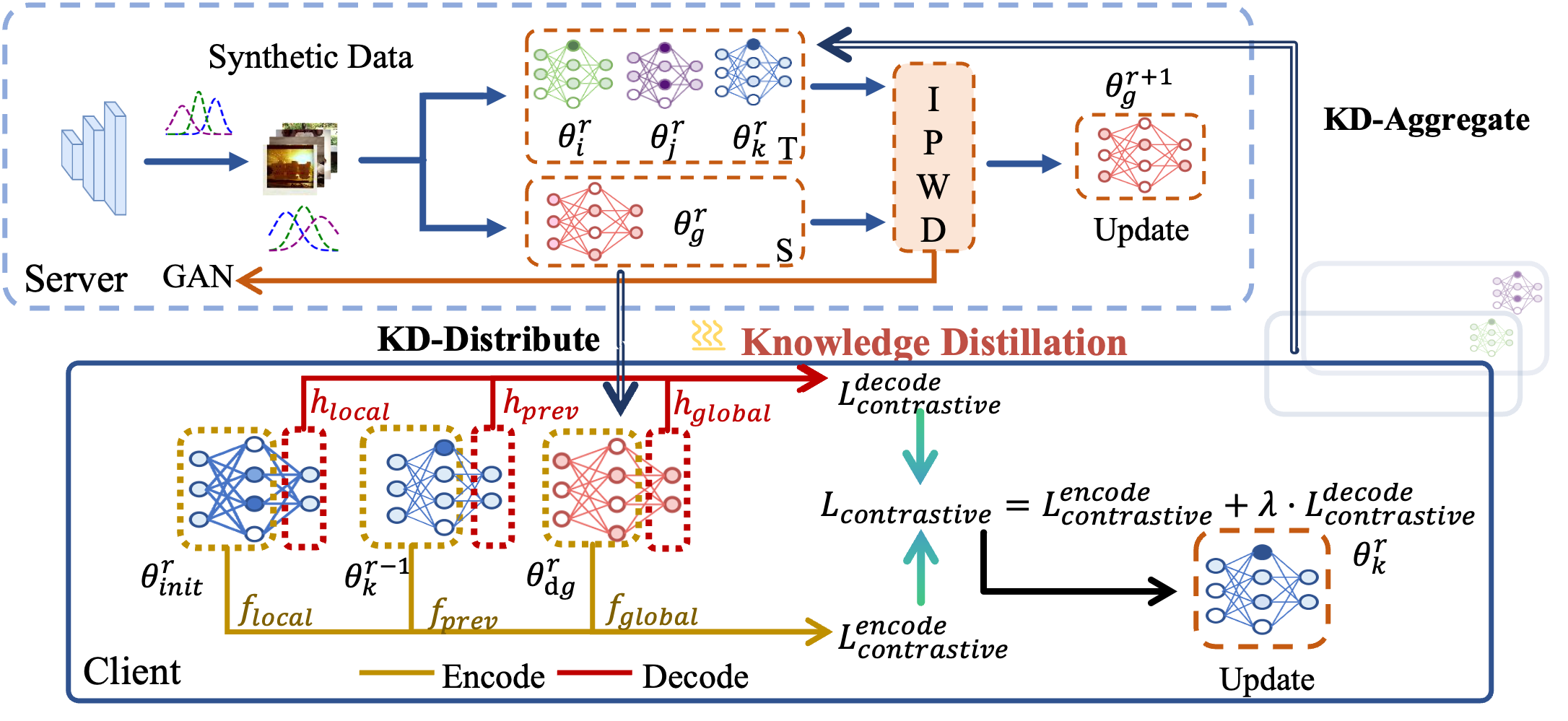} 
	\caption{HFedCKD framework, which consists of the following steps in each communication round: 1.initialize the global model $\theta_{g}^{r}$; 2.distribute $\theta_{gd}^{r}$ to all participating clients via knowledge transfer; 3.initialize the local model $\theta_{init}^{r}$; 4.decompose $\theta_{init}^{r}$ into a feature encode and a decode, and conduct reversed comparative learning with $\theta_{gd}^{r}$ and the historical local model $\theta_{k}^{r-1}$, obtaining the updated local model $\theta_{k}^r$; 5.perform unsupervised knowledge distillation based on \textit{IPWD} to integrate $\theta_{i}^{r}, \theta_{j}^{r}, \theta_{k}^{r}$ with  $\theta_{g}^{r}$, obtaining the updated global model $\theta_{g}^{r+1}$.}
        \label{fig:Figure 1}
\end{figure*}

Assume that $p_i$ represents the proportion of client $i$ data in the overall data distribution:

\begin{equation}
p_i = \frac{|D_i|}{\sum_{j=1}^N |D_j|}.
\end{equation}

Due to the absence of non-participating clients, the sampling distribution of the objective function changes from the data distribution of all clients to the data distribution of participating clients:

\begin{equation}
F_t(w_{\text{global}}) = \sum_{i \in \mathcal{C}_t} \frac{p_i}{\sum_{j \in \mathcal{C}_t} p_j} \mathcal{L}_{\text{KD}, i}.
\end{equation}

The impact of low participation rate is quantified as an error term, and the difference between the ideal objective function and the actual objective function is:

\begin{equation}
\Delta F_t = F(w_{\text{global}}) - F_t(w_{\text{global}}),
\end{equation}

Specifically, the loss term of non-participating clients under low participation rate is ignored:

\begin{equation}
\Delta F_t = \frac{1}{N} \sum_{i \notin \mathcal{C}_t} \mathcal{L}_{\text{KD}, i} - \frac{1}{N} \sum_{i \in \mathcal{C}_t} \left(1 - \frac{N}{|\mathcal{C}_t|} \right) \mathcal{L}_{\text{KD}, i}.
\end{equation}

The loss weight of participating clients is amplified, which will cause the global model update direction to be biased towards the local data distribution of some participating clients, away from the global optimization goal. The gradient size of each round of update may be extremely affected by the data distribution of individual clients under low participation rate, resulting in instability of the global model. As the local model deviates, the model update of non-participating clients may deviate from the global direction due to failure to benefit from global updates.

\section{Framework Overview}
Conduct a theoretical analysis of the impact of low participation rates in heterogeneous environments. As discussed in the previous chapter, the real challenges in FL are mainly low participation rate deviations under heterogeneous models and data. We propose a heterogeneous federated knowledge transfer framework HFedCKD based on bidirectional contrastive learning of IPWD, adopting the following strategy:




\subsection{Data-free knowledge distillation based on IPWD}
Enhance global representation capabilities under low participation rates. HFedCKD introduces a dynamic weighting mechanism based on participation frequency to give higher weight to clients with low frequency participation and balance the global deviation caused by uneven data distribution , effectively makes up for the problem of missing data for non-participating clients, and the distribution of pseudo samples is closer to the real data distribution.





Firstly, we quantify the contribution of each client to the global model by evaluating the relationship between the global pseudo-sample distribution and each client's local data distribution. Based on the IPWD mechanism, we define a corresponding weight \( w_i \) for each client, and the detailed definition and derivation process can be found in the appendix.

\begin{table*}[t!]
    \centering
    \resizebox{\textwidth}{!}{
    \begin{tabular}{cccccccc}
        \toprule
        \textbf{Dataset} & \textbf{Method}    & \textbf{S@10 } & \textbf{S@20} & \textbf{S@50} & \textbf{S@100 } & \textbf{S@200 } & \textbf{S@500 } \\  
        \midrule
        \multirow{3}{*}{\centering Fashion MNIST\cite{xiao2017fashionmnistnovelimagedataset}} 
                             & HeteroFL \cite{diao2021heterofl}          & \textbf{92.39} & \textbf{92.00} & \textbf{90.54} & 88.06 & 87.29 & 86.74 \\ 
                             & FedRolex \cite{alam2022fedrolex}          & 91.76 & 90.14 & 89.08 & \textbf{89.56} & \textbf{88.39} & \textbf{87.43} \\ 
                             & DFRD \cite{10.5555/3666122.3666906}             & 92.13 & 89.95 & 88.01 & 87.73 & 85.23 & 81.21 \\ 
        \midrule
        \multirow{3}{*}{\centering CIFAR100\cite{Krizhevsky09learningmultiple}} 
                             & HeteroFL          & \textbf{38.30} & \textbf{35.03} & 33.42 & \textbf{31.18} & \textbf{29.32} & \textbf{24.41} \\ 
                             & FedRolex          & 34.70 & 34.10 & \textbf{33.59} & 30.14 & 24.78 & 22.22 \\ 
                             & DFRD              & 24.39 & 27.50 & 26.59 & 20.49 & 14.36 & 10.49 \\ 
        \midrule
        \multirow{3}{*}{\centering Tiny-ImageNet\cite{tiny-imagenet}} 
                             & HeteroFL          & 14.52 & 12.54 & 11.10 & 10.80 & 10.61 & 10.69 \\ 
                             & FedRolex          &  \textbf{18.33} & \textbf{18.22} & \textbf{16.72} & \textbf{15.21} & 12.80 & \textbf{11.35} \\ 
                             & DFRD              & 17.53 & 17.06 & 14.98 & 13.53 & \textbf{13.64} & 10.58 \\ 
        \bottomrule              
    \end{tabular}
        }
    \caption{The performance of the limited model heterogeneous FL method compared on three image datasets Fashion MNIST, CIFAR100, and Tiny-ImageNet datasets for image classification tasks  (\%).}
    \label{tab:fashion_mnist_comparison}
\end{table*}

During the knowledge distillation process, the server computes the knowledge transfer loss between the global model and each client model using the generated pseudo-samples \( \{x_j\}_{j=1}^{M} \). Specifically, for each pseudo-sample \( x_j \), we define the prediction distribution of the global model as \( p_{\text{student}}(\mathbf{y}|\mathbf{x}_j) \) and the prediction distribution of the \( k \)-th client model as \( p_{\text{teacher}}^{(k)}(\mathbf{y}|\mathbf{x}_j) \).To further account for the importance of pseudo-samples in the knowledge distillation process, we introduce sample weights \( \gamma_j \),  thereby the weighted KL divergence loss \( \mathcal{L}_{\text{KL}} \) is defined as:

\begin{equation}
    \mathcal{L}_{\text{KL}} = \sum_{k=1}^{K} w_k \sum_{j=1}^{M} \gamma_j \cdot \text{KL}\left( p_{\text{student}}(\mathbf{y}|\mathbf{x}_j) \| p_{\text{teacher}}^{(k)}(\mathbf{y}|\mathbf{x}_j) \right),
\end{equation}


Here, \( \gamma_j \) is dynamically adjusted by the IPWD mechanism based on the sample's confidence and category distribution, specifically defined as:

\begin{equation}
    \gamma_j = \frac{1}{1 + e^{-\lambda (s_j - \theta)}},
\end{equation}
where \( s_j \) is the confidence score of sample \( x_j \), \( \lambda \) is the slope parameter, and \( \theta \) is the threshold.

Finally, by minimizing the weighted KL divergence loss, we update the global model parameters \( \theta_{\text{global}} \) as follows:

\begin{equation}
    \theta_{\text{global}} \leftarrow \theta_{\text{global}} - \eta \nabla_{\theta_{\text{global}}} \mathcal{L}_{\text{KL}},
\end{equation}
where \( \eta \) is the learning rate.

Furthermore, to ensure comprehensive coverage of the data distribution, the system generates an appropriate number of pseudo-samples for any detected missing categories. 

\begin{equation}
    \{(\mathbf{x}_i, y_i)\}_{\text{miss}} = \text{Generator}(\mathbf{z}_{\text{miss}}, y_{\text{miss}}),
\end{equation}
where \( \mathbf{z}_{\text{miss}} \) is the random noise generated for missing categories, and \( y_{\text{miss}} \) denotes the pseudo-labels for   missing categories.

\subsection{Encode/decode hierarchical bidirectional contrastive learning}
HFedCKD uses bidirectional contrastive learning to solve the deviation problem from two levels: global feature consistency and local personalized optimization: it can not only improve the consistency of global features, but also retain personalization for each client decision-making capabilities .

\paragraph{Encode-Global Alignment}: The local model is divided into two components: a feature extractor (Encoder) and a classifier (Classifier). The feature extractor is responsible for extracting high-level features , while the classifier makes decisions based on these features. Through contrastive learning, we optimize the local feature extractor to align the extracted features with those of the global model's feature space.

Specifically, we define the local feature extractor as \( f_{\text{local}} \), the global feature extractor as \( f_{\text{global}} \), and the historical local feature extractor as \( f_{\text{prev}} \). Utilizing the generated pseudo-samples \( \{x_j\}_{j=1}^{M} \), we extract feature vectors:
\begin{equation}
    z_{\text{l}}^{(j)} = f_{\text{local}}(x_j),  z_{\text{g}}^{(j)} = f_{\text{global}}(x_j),  z_{\text{p}}^{(j)} = f_{\text{prev}}(x_j)
\end{equation}

\begin{table*}[t!]
    \centering
    \resizebox{\textwidth}{!}{
    \begin{tabular}{cccccccc}
        \toprule
        \textbf{Dataset} & \textbf{Method}    & \textbf{S@10 } & \textbf{S@20} & \textbf{S@50} & \textbf{S@100 } & \textbf{S@200 } & \textbf{S@500 } \\  
        \midrule
        \multirow{7}{*}{\centering Fashion MNIST} 
                             & Baseline          & 92.50 & 90.08 & 87.20 & 84.84 & 77.51 & 65.20 \\ 
                             & FedMD  \cite{li2019fedmd}           & 88.78 & 92.40 & 92.37 & 84.52 & 81.36 & 79.74 \\ 
                             & FedGen \cite{venkateswaran2023fedgen}           & 88.76     & 88.60     & 86.98     & 86.89     & 85.29     & 80.86     \\ 
                             & FedAlign \cite{mendieta2022local}         & 89.61     & 89.52     & 88.73     & 86.49     & 82.34     & 71.06     \\ 
                             \cline{2-8}  
                             & HFedCKD (w/o. BCL) & 92.41 & 90.07 & 86.33 & 84.74 & 81.75 & 75.22 \\ 
                             & HFedCKD (w/o. IPWD) & 92.53 & 90.38 & 89.91 & 88.21 & 85.96 & 78.78 \\ 
                             & HFedCKD           & \textbf{92.73} & \textbf{91.36} & \textbf{90.82} & \textbf{89.40} & \textbf{88.22} & \textbf{81.70} \\ 
        \midrule
        \multirow{7}{*}{\centering CIFAR100} 
                             & Baseline          & 36.81 & 35.76 & 31.93 & 29.22 & 27.35 & 21.10 \\ 
                             & FedMD             & 27.33 & 27.99 & 28.58 & 23.15 & 14.35 & 11.45 \\ 
                             & FedGen            & 27.07     & 25.89     & 22.14     & 19.24     & 18.66     & 16.49     \\ 
                             & FedAlign          & \textbf{48.47} & \textbf{44.75}    & 37.31     & 32.18     & 27.68     & 19.97     \\ 
                             \cline{2-8} 
                             & HFedCKD (w/o. BCL) & 25.70 & 26.79 & 28.44 & 27.87 & 27.17 & 24.00 \\ 
                             & HFedCKD (w/o. IPWD) & 42.52 & 36.56 & 40.04 & 39.60 & 32.74 & 26.37 \\ 
                             & HFedCKD           & 42.55 & 41.87 & \textbf{40.69} & \textbf{39.83} & \textbf{33.56} & \textbf{26.86} \\ 
        \midrule
        \multirow{7}{*}{\centering Tiny-ImageNet} 
                             & Baseline          & 29.70 & 28.15 & 27.85 & 23.01 & 20.79 & 16.15 \\ 
                             & FedMD             & 19.98  & 15.43 & 11.23 & 10.18 & 10.50  & 9.76     \\ 
                             & FedGen            & 20.13     & 18.22     & 13.59     & 11.54     & 10.65     & 9.48     \\ 
                             & FedAlign          & \textbf{33.60}  & \textbf{32.10} & 28.09  &21.78   &16.64   &10.70    \\ 
                             \cline{2-8} 
                             & HFedCKD (w/o. BCL) & 20.04 & 20.51 & 22.05 & 21.53 & 21.61 & 18.45 \\ 
                             & HFedCKD (w/o. IPWD) & 28.33 & 29.67 & 28.82 & 25.31 & 22.62 & 19.63 \\ 
                             & HFedCKD           & 28.57 & 29.74 & \textbf{28.85} & \textbf{25.81} & \textbf{23.05} & \textbf{20.01} \\ 
        \bottomrule              
    \end{tabular}
        }
    \caption{The performance of unlimited heterogeneous FL methods compared on three image datasets Fashion MNIST, CIFAR100, and Tiny-ImageNet datasets for image classification tasks (\%).}
    \label{tab:fashion_mnist_comparison_hetero}
\end{table*}

By maximizing the similarity between local features and global features while minimizing the similarity between local features and historical local features, we define the encoding contrastive loss \( L_{\text{contrastive}}^{\text{encode}} \) as:
\begin{equation}
a_j = \exp\left( \frac{\text{cos}(z_{\text{l}}^{(j)}, z_{\text{g}}^{(j)})}{\tau} \right),b_k = \exp\left( \frac{\text{cos}(z_{\text{l}}^{(j)}, z_{\text{p}}^{(k)})}{\tau} \right)
\end{equation}
\begin{equation}
    L_{\text{contrastive}}^{\text{encode}} = \frac{1}{M} \sum_{j=1}^{M}
    \left[  \log \left( \frac{a_j}{a_j + \sum_{k=1}^{K} b_k} \right) \right]
\end{equation}

where \( \text{cos}(\cdot, \cdot) \) denotes cosine similarity, \( \tau \) is the temperature coefficient, and \( K \) is the number of negative samples.


\paragraph{Decode-History Alignment}: The classifier component retains the local model's personalized decision-making capability. By aligning with the historical local model's features, it ensures that the local model does not deviate from its data characteristics during continuous updates.

Similar to the feature extractor alignment process, the classifier optimizes through contrastive learning to reference the historical model's feature representations during classification decisions.
Specifically, we define the classifier as \( \text{C}(\cdot) \), with the local, global, and historical classification vectors represented as:
\begin{equation}
    h_{\text{l}}^{(j)} = C_{\text{local}}(z_j), \quad h_{\text{g}}^{(j)} = C_{\text{global}}(z_j), \quad h_{\text{p}}^{(j)} = C_{\text{prev}}(z_j)
\end{equation}

The classification contrastive loss \( L_{\text{contrastive}}^{\text{decode}} \) is defined as:
\begin{equation}
c_j = \exp\left( \frac{\text{cos}(h_{\text{l}}^{(j)}, h_{\text{p}}^{(j)})}{\tau} \right),d_k = \exp\left( \frac{\text{cos}(h_{\text{l}}^{(j)}, h_{\text{g}}^{(k)})}{\tau} \right)
\end{equation}
\begin{equation}
    L_{\text{contrastive}}^{\text{decode}} = \frac{1}{M} \sum_{j=1}^{M}
    \left[ - \log \left( \frac{c_j}{c_j + \sum_{k=1}^{K} d_k} \right) \right]
\end{equation}
where \( \text{cos}(\cdot, \cdot) \) denotes cosine similarity, \( \tau \) is the temperature coefficient, and \( K \) is the number of negative samples.
\paragraph{Comprehensive Optimization of the Loss Function}

We combine the contrastive losses of the feature extractor and the classifier to form the overall contrastive learning loss. To further enhance the effectiveness of contrastive learning, we introduce a multi-layer contrastive learning strategy.

\begin{equation}
    L_{\text{contrastive}} = \sum_{l=1}^{L} \lambda_l \left( L_{\text{contrastive}, l}^{\text{encode}} + \lambda \cdot L_{\text{contrastive}, l}^{\text{decode}} \right)
\end{equation}
where \( \lambda \) is a weighting coefficient used to balance the importance of the two loss terms.\( L_{\text{contrastive}, l}^{\text{encode}} \) and \( L_{\text{contrastive}, l}^{\text{decode}} \) are the encoding and classification contrastive losses at the \( l \)-th layer, respectively, and \( \lambda_l \) is the corresponding layer weighting coefficient.




Through the above comprehensive contrastive learning loss, we optimize the local model's feature space to align with the global model, while preserving its personalized decision-making capability. Furthermore, the multi-layer contrastive strategy helps capture feature representations at different levels, thereby enhancing the model's adaptability.

Ultimately, the local model's update objective function combines the knowledge distillation loss and the contrastive learning loss, defined as:

\begin{equation}
    \mathcal{L}_{\text{total}} = \mathcal{L}_{\text{KD}} + \gamma \cdot L_{\text{contrastive}},
\end{equation}
where \( \gamma \) is a weighting coefficient used to balance the influence of the knowledge distillation loss and the contrastive learning loss.

\begin{figure*}[t!]
	\centering
        \includegraphics[height=5cm]{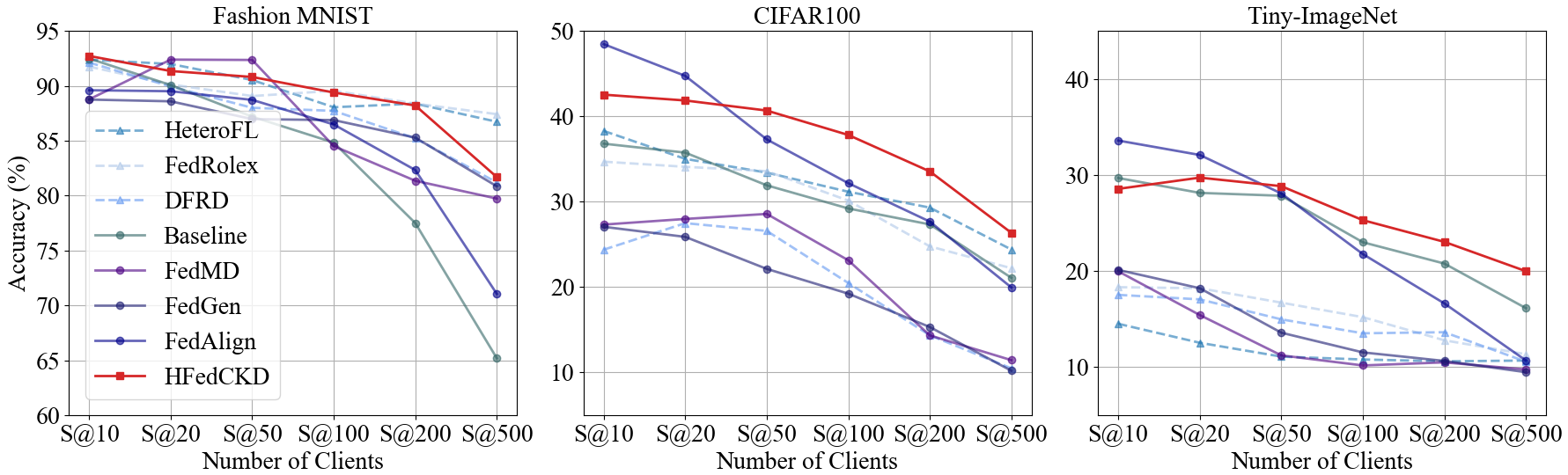} 
	\caption{The performance of the limited model and unlimited heterogeneous FL method, Intuitively and synthetically.}
        \label{fig:Figure 1}
\end{figure*}


\section{Experiment}

\subsection{Experiments Setting and Evaluation}

\textbf{Datasets and experimental settings.} Refer to the appendix for datasets and experimental Settings.

\textbf{Data heterogeneity}. The Dirichlet process Dir($\alpha$) is used to split the training set of each dataset to generate non-independent and identically distributed data between clients, so as to allocate local training data to each client. $\alpha$ is a centralized parameter. Smaller $\alpha$ corresponds to stronger data heterogeneity. The default $\alpha$ is set to 0.1 to simulate a strict non-independent and identically distributed scenario.

\textbf{Model heterogeneity}. The initial global model is ResNet18. We divide the comparative experiments into two categories. The first category can only achieve model heterogeneity in a narrow sense, that is, to allow the client model to share the architecture through scaling. We set up three different clients based on Heterofl. Model capacity$\beta$={1,1/2,1/4}, corresponding to the proportion of client models is {40\%,30\%,30\%}. The second category is generalized model heterogeneity. The client models are initialized as FedAvgMLP, LeNet, and CifarNet respectively, and the corresponding client models account for {40\%,30\%,30\%}.

\textbf{Experimental evaluation}. We conduct image classification experiments on three classic image datasets Fashion MNIST, Cifar100, and Tiny-ImageNet to evaluate the effectiveness of the proposed HFL framework. The experiments are divided into limited model heterogeneity and full model heterogeneity. In Hetero, the personalized accuracy of other HFL algorithms (such as FedMD , FedGen, FedAlign, etc.) under the same experimental settings is compared, and all experimental results are presented as the average of multiple experiments. We use a client-level evaluation strategy to analyze the client model performance of different methods to verify the advantages of HFedCKD, which shows that HFedCKD has excellent generalization and stability \cite{li2019fedmd,venkateswaran2023fedgen,mendieta2022local}.
\subsection{Experiment Results}
\textbf{Performance comparison experiment with other model heterogeneous methods.} The results with the best performance are marked in bold. Experimental results show that in scenarios where both data heterogeneity and model heterogeneity exist, HFedCKD always outperforms most baseline methods in image classification tasks. Especially as the data scale increases and the client participation rate decreases, the superior performance of HFedCKD becomes more and more obvious, demonstrating excellent adaptability and robustness in complex environments.

\textbf{Ablation Studies. }In order to evaluate the contribution of IPWD component and HCL component in our proposed HFedCKD framework, we conducted ablation experiments to verify the effectiveness of HFedCKD framework design. We removed two key components of the framework and obtained the HFedCKD (w/o. IPWD) model and the HFedCKD (w/o. BCL) model.
\begin{itemize}
    \item HFedCKD (w/o. BCL): We removed the differential contrast learning mechanism from HFedCKD, and the client obtained the global model features through knowledge distillation for local training.
    \item HFedCKD (w/o. IPWD): We disabled IPWD to handle the knowledge distillation link and adopted the general communication strategy. After aggregating the client models, the pre-trained model was knowledge distilled on the global update model.
\end{itemize}
During the experiment, we found that the HFedCKD (w/o. IPWD) model performed unstable on some datasets generated based on random seeds, which was specifically manifested in the gradient explosion phenomenon during the S@200 and S@500 experiments. . The analysis reason is that the insufficient number of clients participating leads to insufficient information transmission during local update of the model. The lack of effective IPWD processing mechanism prevents the heterogeneity between clients from being fully considered, which in turn causes training interruption. HFedCKD(w/o. BCL) performance is more stable. The experimental performance and experimental results verify the effectiveness of the key components of the HFedCKD framework.

\begin{table}[h]
    \centering
    \begin{tabular}{ccccc}
         \toprule
        \textbf{Dataset} & \textbf{Jr=1} & \textbf{2/3} & \textbf{1/3} & \textbf{1/9} \\ 
        \midrule
        UCI-HAR &90.52  &90.18  &88.89  &87.77  \\ 
        PAMAP2  &94.95  &94ou.83  &94.75  &94.69  \\ 
        \bottomrule
    \end{tabular}
    \caption{Performance of HFedCKD retrieval tasks based on UCI-HAR and PAMAP2 datasets is accompanied with different participation rates jr (\%)}
    \label{tab:dataset_comparison}
\end{table}

\textbf{HFedCKD multimodal generalization performance experiment.}
HFedCKD can also effectively support IoT datasets and shows good robustness. We conducted multiple sets of experiments on the IoT dataset PAMAP2 and UCI-HAR \cite{DBLPconfpetraReissS12,human_activity_recognition_using_smartphones_240} respectively. The experimental setting covers participation rates jr = \{1,2/3,1/3,1/9\}. Experimental results show that although the model performance of HFedCKD decreases slightly on the UCI-HAR and PAMAP2 datasets as the participation rate decreases, the overall performance is still excellent. And the performance is more stable on PAMAP2.

\section{Conclusion}
We propose a novel systematic HFL framework based on data-free knowledge distillation and bidirectional contrastive learning (HFedCKD), which aims to solve the performance degradation problem caused by low participation rate in FL under heterogeneous data and model scenarios. By introducing the IPWD strategy, HFedCKD dynamically adjusts the weight allocation of clients, enhances the contribution of low-frequency participating clients in global model training, and significantly alleviates the distribution deviation of generator pseudo-samples and feature missing problems. At the same time, by setting hierarchical contrastive learning, HFedCKD realizes the feature alignment between the feature extractor and the global model, and the personalized optimization path protection of the classifier, which improves the feature expression ability of the global model and the decision-making ability of the local model. Experiments in several different scenarios verify the excellent generalization and robustness of HFedCKD. Compared with the existing methods, HFedCKD further enhances the adaptability of the model in heterogeneous scenarios, can effectively combat label bias and feature drift in low participation rate Settings, and realizes efficient collaboration of diverse clients.

\bibliographystyle{named}
\bibliography{ijcai24}

\begin{thebibliography}{}

\bibitem[\protect\citeauthoryear{Alam \bgroup \em et al.\egroup }{2022}]{alam2022fedrolex}
Samiul Alam, Luyang Liu, Ming Yan, and Mi~Zhang.
\newblock Fedrolex: Model-heterogeneous federated learning with rolling sub-model extraction.
\newblock {\em Advances in neural information processing systems}, 35:29677--29690, 2022.

\bibitem[\protect\citeauthoryear{Beitollahi \bgroup \em et al.\egroup }{2022}]{10019204}
Mahdi Beitollahi, Mingrui Liu, and Ning Lu.
\newblock Dsfl: Dynamic sparsification for federated learning.
\newblock In {\em 2022 5th International Conference on Communications, Signal Processing, and their Applications (ICCSPA)}, pages 1--6, 2022.

\bibitem[\protect\citeauthoryear{Chen \bgroup \em et al.\egroup }{2023}]{chen2023towards}
Jinqian Chen, Jihua Zhu, and Qinghai Zheng.
\newblock Towards fast and stable federated learning: Confronting heterogeneity via knowledge anchor.
\newblock In {\em Proceedings of the 31st ACM International Conference on Multimedia}, pages 8697--8706, 2023.

\bibitem[\protect\citeauthoryear{Chen \bgroup \em et al.\egroup }{2024}]{conf/aaai/ChenZZLT24}
Jinqian Chen, Jihua Zhu, Qinghai Zheng, Zhongyu Li, and Zhiqiang Tian.
\newblock Watch your head: Assembling projection heads to save the reliability of federated models.
\newblock In Michael~J. Wooldridge, Jennifer~G. Dy, and Sriraam Natarajan, editors, {\em AAAI}, pages 11329--11337. AAAI Press, 2024.

\bibitem[\protect\citeauthoryear{Chen \bgroup \em et al.\egroup }{2025}]{10.24963/ijcai.2024/419}
Jinqian Chen, Haoyu Tang, Junhao Cheng, Ming Yan, Ji~Zhang, Mingzhu Xu, Yupeng Hu, and Liqiang Nie.
\newblock Breaking barriers of system heterogeneity: straggler-tolerant multimodal federated learning via knowledge distillation.
\newblock In {\em Proceedings of the Thirty-Third International Joint Conference on Artificial Intelligence}, IJCAI '24, 2025.

\bibitem[\protect\citeauthoryear{Collins \bgroup \em et al.\egroup }{2021}]{collins2021exploiting}
Liam Collins, Hamed Hassani, Aryan Mokhtari, and Sanjay Shakkottai.
\newblock Exploiting shared representations for personalized federated learning.
\newblock In {\em International conference on machine learning}, pages 2089--2099. PMLR, 2021.

\bibitem[\protect\citeauthoryear{Diao \bgroup \em et al.\egroup }{2021}]{diao2021heterofl}
Enmao Diao, Jie Ding, and Vahid Tarokh.
\newblock Hetero{\{}fl{\}}: Computation and communication efficient federated learning for heterogeneous clients.
\newblock In {\em International Conference on Learning Representations}, 2021.

\bibitem[\protect\citeauthoryear{Feng \bgroup \em et al.\egroup }{2023}]{feng2023hierarchical}
Xin Feng, Yifeng Xu, Guangming Lu, and Wenjie Pei.
\newblock Hierarchical contrastive learning for pattern-generalizable image corruption detection.
\newblock In {\em Proceedings of the IEEE/CVF International Conference on Computer Vision}, pages 12076--12085, 2023.

\bibitem[\protect\citeauthoryear{Han \bgroup \em et al.\egroup }{2024}]{10.1109/JSAC.2024.3431516}
Pengchao Han, Xingyan Shi, and Jianwei Huang.
\newblock Fedal: Black-box federated knowledge distillation enabled by adversarial learning.
\newblock {\em IEEE J.Sel. A. Commun.}, 42(11):3064–3077, July 2024.

\bibitem[\protect\citeauthoryear{He \bgroup \em et al.\egroup }{2020}]{he2020group}
Chaoyang He, Murali Annavaram, and Salman Avestimehr.
\newblock Group knowledge transfer: Federated learning of large cnns at the edge.
\newblock {\em Advances in Neural Information Processing Systems}, 33:14068--14080, 2020.

\bibitem[\protect\citeauthoryear{Horvath \bgroup \em et al.\egroup }{2021}]{horvath2021fjord}
Samuel Horvath, Stefanos Laskaridis, Mario Almeida, Ilias Leontiadis, Stylianos Venieris, and Nicholas Lane.
\newblock Fjord: Fair and accurate federated learning under heterogeneous targets with ordered dropout.
\newblock {\em Advances in Neural Information Processing Systems}, 34:12876--12889, 2021.

\bibitem[\protect\citeauthoryear{Jing \bgroup \em et al.\egroup }{2025}]{10.5555/3692070.3692967}
Shusen Jing, Anlan Yu, Shuai Zhang, and Songyang Zhang.
\newblock Fedsc: provable federated self-supervised learning with spectral contrastive objective over non-i.i.d. data.
\newblock In {\em Proceedings of the 41st International Conference on Machine Learning}, ICML'24. JMLR.org, 2025.

\bibitem[\protect\citeauthoryear{Krizhevsky}{2009}]{Krizhevsky09learningmultiple}
Alex Krizhevsky.
\newblock Learning multiple layers of features from tiny images.
\newblock Technical report, 2009.

\bibitem[\protect\citeauthoryear{Lee \bgroup \em et al.\egroup }{2022}]{lee2022preservation}
Gihun Lee, Minchan Jeong, Yongjin Shin, Sangmin Bae, and Se-Young Yun.
\newblock Preservation of the global knowledge by not-true distillation in federated learning.
\newblock {\em Advances in Neural Information Processing Systems}, 35:38461--38474, 2022.

\bibitem[\protect\citeauthoryear{Li and Wang}{2019}]{li2019fedmd}
Daliang Li and Junpu Wang.
\newblock Fedmd: Heterogenous federated learning via model distillation.
\newblock 2019.

\bibitem[\protect\citeauthoryear{Li \bgroup \em et al.\egroup }{2021}]{li2021model}
Qinbin Li, Bingsheng He, and Dawn Song.
\newblock Model-contrastive federated learning.
\newblock In {\em Proceedings of the IEEE/CVF conference on computer vision and pattern recognition}, pages 10713--10722, 2021.

\bibitem[\protect\citeauthoryear{Liu \bgroup \em et al.\egroup }{2020}]{Liu2020}
Yuan Liu, Zhengpeng Ai, Shuai Sun, Shuangfeng Zhang, Zelei Liu, and Han Yu.
\newblock {\em FedCoin: A Peer-to-Peer Payment System for Federated Learning}, pages 125--138.
\newblock Springer International Publishing, Cham, 2020.

\bibitem[\protect\citeauthoryear{Liu \bgroup \em et al.\egroup }{2024}]{10.1145/3639706}
Jing Liu, Litao Shang, Yuting Su, Weizhi Nie, Xin Wen, and Anan Liu.
\newblock Privacy-preserving multi-source cross-domain recommendation based on knowledge graph.
\newblock {\em ACM Trans. Multimedia Comput. Commun. Appl.}, 20(5), February 2024.

\bibitem[\protect\citeauthoryear{Luo \bgroup \em et al.\egroup }{2024}]{10.5555/3666122.3666906}
Kangyang Luo, Shuai Wang, Yexuan Fu, Xiang Li, Yunshi Lan, and Ming Gao.
\newblock Dfrd: data-free robustness distillation for heterogeneous federated learning.
\newblock In {\em Proceedings of the 37th International Conference on Neural Information Processing Systems}, NIPS '23, Red Hook, NY, USA, 2024. Curran Associates Inc.

\bibitem[\protect\citeauthoryear{McMahan \bgroup \em et al.\egroup }{2017}]{mcmahan2017communication}
Brendan McMahan, Eider Moore, Daniel Ramage, Seth Hampson, and Blaise~Aguera y~Arcas.
\newblock Communication-efficient learning of deep networks from decentralized data.
\newblock In {\em Artificial intelligence and statistics}, pages 1273--1282. PMLR, 2017.

\bibitem[\protect\citeauthoryear{Mendieta \bgroup \em et al.\egroup }{2022}]{mendieta2022local}
Matias Mendieta, Taojiannan Yang, Pu~Wang, Minwoo Lee, Zhengming Ding, and Chen Chen.
\newblock Local learning matters: Rethinking data heterogeneity in federated learning.
\newblock In {\em Proceedings of the IEEE/CVF Conference on Computer Vision and Pattern Recognition}, pages 8397--8406, 2022.

\bibitem[\protect\citeauthoryear{mnmoustafa and Ali}{2017}]{tiny-imagenet}
mnmoustafa and Mohammed Ali.
\newblock Tiny imagenet.
\newblock \url{https://kaggle.com/competitions/tiny-imagenet}, 2017.
\newblock Kaggle.

\bibitem[\protect\citeauthoryear{Niu \bgroup \em et al.\egroup }{2022}]{niu2022respecting}
Yulei Niu, Long Chen, Chang Zhou, and Hanwang Zhang.
\newblock Respecting transfer gap in knowledge distillation.
\newblock {\em Advances in Neural Information Processing Systems}, 35:21933--21947, 2022.

\bibitem[\protect\citeauthoryear{Peng \bgroup \em et al.\egroup }{2023}]{10.1007/978-3-031-39698-4_23}
Chao Peng, Yiming Guo, Yao Chen, Qilin Rui, Zhengfeng Yang, and Chenyang Xu.
\newblock Fedgm: Heterogeneous federated learning via\&nbsp;generative learning and\&nbsp;mutual distillation.
\newblock In {\em Euro-Par 2023: Parallel Processing: 29th International Conference on Parallel and Distributed Computing, Limassol, Cyprus, August 28 – September 1, 2023, Proceedings}, page 339–351, Berlin, Heidelberg, 2023. Springer-Verlag.

\bibitem[\protect\citeauthoryear{Peng \bgroup \em et al.\egroup }{2024}]{Peng2024FedPFTFP}
Zhaopeng Peng, Xiaoliang Fan, Yufan Chen, Zheng Wang, Shirui Pan, Chenglu Wen, Ruisheng Zhang, and Cheng Wang.
\newblock Fedpft: Federated proxy fine-tuning of foundation models.
\newblock In {\em International Joint Conference on Artificial Intelligence}, 2024.

\bibitem[\protect\citeauthoryear{Reddi \bgroup \em et al.\egroup }{2021}]{reddi2021adaptive}
Sashank~J. Reddi, Zachary Charles, Manzil Zaheer, Zachary Garrett, Keith Rush, Jakub Kone{\v{c}}n{\'y}, Sanjiv Kumar, and Hugh~Brendan McMahan.
\newblock Adaptive federated optimization.
\newblock In {\em International Conference on Learning Representations}, 2021.

\bibitem[\protect\citeauthoryear{Reiss and Stricker}{2012}]{DBLPconfpetraReissS12}
Attila Reiss and Didier Stricker.
\newblock Creating and benchmarking a new dataset for physical activity monitoring.
\newblock In Fillia Makedon, editor, {\em The 5th International Conference on PErvasive Technologies Related to Assistive Environments, {PETRA} 2012, Heraklion, Crete, Greece, June 6-9, 2012}, page~40. {ACM}, 2012.

\bibitem[\protect\citeauthoryear{Reyes-Ortiz \bgroup \em et al.\egroup }{2013}]{human_activity_recognition_using_smartphones_240}
Jorge Reyes-Ortiz, Davide Anguita, Alessandro Ghio, Luca Oneto, and Xavier Parra.
\newblock Human activity recognition using smartphones.
\newblock UCI Machine Learning Repository, 2013.
\newblock DOI: <https://doi.org/10.24432/C54S4K>.

\bibitem[\protect\citeauthoryear{Shi \bgroup \em et al.\egroup }{2020}]{Shi2020JointDS}
Wenqi Shi, Sheng Zhou, Zhisheng Niu, Miao Jiang, and Lu~Geng.
\newblock Joint device scheduling and resource allocation for latency constrained wireless federated learning.
\newblock {\em IEEE Transactions on Wireless Communications}, 20:453--467, 2020.

\bibitem[\protect\citeauthoryear{Tu \bgroup \em et al.\egroup }{2024}]{10.1145/3638252}
Jingke Tu, Jiaming Huang, Lei Yang, and Wanyu Lin.
\newblock Personalized federated learning with layer-wise feature transformation via meta-learning.
\newblock {\em ACM Trans. Knowl. Discov. Data}, 18(4), February 2024.

\bibitem[\protect\citeauthoryear{Usmanova \bgroup \em et al.\egroup }{2022}]{usmanova2022federated}
Anastasiia Usmanova, Fran{\c{c}}ois Portet, Philippe Lalanda, and German Vega.
\newblock Federated continual learning through distillation in pervasive computing.
\newblock In {\em 2022 IEEE International Conference on Smart Computing (SMARTCOMP)}, pages 86--91. IEEE, 2022.

\bibitem[\protect\citeauthoryear{Venkateswaran \bgroup \em et al.\egroup }{2023}]{venkateswaran2023fedgen}
Praveen Venkateswaran, Vatche Isahagian, Vinod Muthusamy, and Nalini Venkatasubramanian.
\newblock Fedgen: Generalizable federated learning for sequential data.
\newblock In {\em 2023 IEEE 16th International Conference on Cloud Computing (CLOUD)}, pages 308--318. IEEE, 2023.

\bibitem[\protect\citeauthoryear{Wu \bgroup \em et al.\egroup }{2021}]{Wu2021FedCGLC}
Yuezhou Wu, Yan Kang, Jiahuan Luo, Yuanqin He, and Qiang Yang.
\newblock Fedcg: Leverage conditional gan for protecting privacy and maintaining competitive performance in federated learning.
\newblock In {\em International Joint Conference on Artificial Intelligence}, 2021.

\bibitem[\protect\citeauthoryear{Wu \bgroup \em et al.\egroup }{2024}]{wu2024exploring}
Zhiyuan Wu, Sheng Sun, Yuwei Wang, Min Liu, Quyang Pan, Junbo Zhang, Zeju Li, and Qingxiang Liu.
\newblock Exploring the distributed knowledge congruence in proxy-data-free federated distillation.
\newblock {\em ACM Transactions on Intelligent Systems and Technology}, 15(2):1--34, 2024.

\bibitem[\protect\citeauthoryear{Xiao \bgroup \em et al.\egroup }{2017}]{xiao2017fashionmnistnovelimagedataset}
Han Xiao, Kashif Rasul, and Roland Vollgraf.
\newblock Fashion-mnist: a novel image dataset for benchmarking machine learning algorithms, 2017.

\bibitem[\protect\citeauthoryear{Yu \bgroup \em et al.\egroup }{2025}]{10.1145/3704323.3704370}
Peng Yu, Yuwen Yuan, Youquan Wang, Wei Shi, Zhihui Shi, and Yue Hua.
\newblock Fedrfd: Heterogeneous federated learning via refining feature-level distillation.
\newblock In {\em Proceedings of the 2024 13th International Conference on Computing and Pattern Recognition}, ICCPR '24, page 287–294, New York, NY, USA, 2025. Association for Computing Machinery.

\bibitem[\protect\citeauthoryear{Zhang \bgroup \em et al.\egroup }{2021}]{zhang2021federated}
Tuo Zhang, Chaoyang He, Tianhao Ma, Lei Gao, Mark Ma, and Salman Avestimehr.
\newblock Federated learning for internet of things.
\newblock In {\em Proceedings of the 19th ACM Conference on Embedded Networked Sensor Systems}, pages 413--419, 2021.

\bibitem[\protect\citeauthoryear{Zhu \bgroup \em et al.\egroup }{2021}]{zhu2021federated}
Hangyu Zhu, Jinjin Xu, Shiqing Liu, and Yaochu Jin.
\newblock Federated learning on non-iid data: A survey.
\newblock {\em Neurocomputing}, 465:371--390, 2021.

\end{thebibliography}

\end{document}